\pdfoutput=1

\documentclass[11pt]{article}

\usepackage[final]{acl}

\usepackage{times}
\usepackage{latexsym}

\usepackage[T1]{fontenc}

\usepackage[utf8]{inputenc}

\usepackage{microtype}

\usepackage{inconsolata}

\usepackage{graphicx}

\usepackage{xspace}
\usepackage{booktabs}
\usepackage{soul}

\title{Cacheback: Speculative Decoding With Nothing But Cache}

\author{
Zhiyao Ma$^*$ \and In Gim$^*$ \and Lin Zhong \\
Yale University \\
\texttt{\{zhiyao.ma,in.gim,lin.zhong\}@yale.edu}
}

\newcommand{\fullname}{Cacheback Decoding\xspace}
\newcommand{\name}{Cacheback\xspace}
\newcommand{\leadlen}{LL\xspace}
\newcommand{\folllen}{FL\xspace}
\newcommand{\leadcap}{LC\xspace}
\newcommand{\follcap}{FC\xspace}
\newcommand{\tdl}{TDL\xspace}
\newcommand{\crt}{CRT\xspace}

\begin{document}
\maketitle

\begingroup
\renewcommand{\thefootnote}{\fnsymbol{footnote}}
\footnotetext[1]{Equal contribution.}
\endgroup

\begin{abstract}
We present \emph{\fullname}, a training-free and model-agnostic speculative decoding method that exploits the locality in language to accelerate Large Language Model (LLM) inference.
\name leverages only Least Recently Used (LRU) cache tables of token n-grams to generate draft sequences.
\name achieves state-of-the-art performance among comparable methods despite its minimalist design, and its simplicity allows easy integration into existing systems.
\name also shows potential for fast adaptation to new domains.
\end{abstract}

\section{Introduction}
\label{sec:introduction}

Cache Language Models (CLMs), notable innovations from the 1990s~\cite{kuhn1990cache}, enhanced the predictive capabilities of n-gram models.
They store recently observed n-grams in a cache.
Using the cache table, they modify the probabilities assigned by the base n-gram model to favor n-grams present in the cache, effectively exploiting the linguistic phenomenon of ``burstiness,'' i.e., the increased likelihood of recently used words reappearing.

With the subsequent rise of Large Language Models (LLMs), whose massive parameterization enables them to capture complex, long-distance contextual patterns, the original purpose of CLMs appears to have been superseded.
However, we re-examine the utility of caching not to improve the intrinsic modeling power of already potent LLMs, but as a surprisingly effective tool for a different objective: accelerating LLM generative inference.

We present \emph{\fullname}, a novel method that repurposes the CLM concept for the modern challenge of Speculative Decoding (SD).
In the SD framework~\cite{leviathan2023fast}, a faster mechanism proposes a sequence of draft tokens, which the LLM then attempts to validate in a single forward pass, potentially accepting multiple tokens at once and thereby reducing overall latency.
\name generates these drafts without auxiliary neural models or complex algorithmic procedures.

\name's drafting mechanism is extremely simple:
It maintains a cache table with the Least Recently Used (LRU) eviction policy.
This table maps a tuple of leading tokens to a set of tuples of immediately following tokens most recently observed after the leading ones in the ongoing generation process or recent context.
\name generates a tree of draft tokens by recursively querying the cache table using the last few tokens in a tree branch as the key and retrieving the follower tokens to grow the tree.
This draft generation step is lightweight, typically executing in microseconds, thus imposing negligible overhead on the decoding loop.

Our empirical evaluations on the SpecBench benchmark~\citep{xia2024unlocking} demonstrate that \name, despite its minimalist design, achieves state-of-the-art performance in wall-clock speedup and token acceptance ratio among comparable baselines that do not require draft model training or model architecture modifications.
The effectiveness of \name suggests avenues for future work, including dynamic cache scaling and rapid domain adaptation for draft generation, a traditional strength of CLMs.

\section{Background and Related Work}
\label{sec:background}

\subsection{Exploiting Locality in Language Modeling}

The principle of locality, referring to the tendency for related words to appear in close proximity, is a universal characteristic of both artificial (e.g., programming languages) and natural languages.
This phenomenon is theoretically grounded in information theory.
As \citet{futrell2015large} posits, if we consider the limitations of human information processing and the constraints of short-term memory, then efficient languages are expected to favor local information structures.
That is, \emph{words that are linked in meaning or usage should occur near each other}.
This inherent locality can be effectively exploited using surprisingly simple mechanisms like caching.
CLMs were pioneering in this regard, integrating n-gram caches to enhance language modeling quality by re-weighting probabilities from a base n-gram model.
Despite their simplicity, CLMs demonstrated significant empirical efficacy, achieving perplexity reductions of 38\% to 50\% in tasks like speech recognition~\citep{jelinek1991dynamic,clarkson1997language}, which led to their popularity in the 1990s.

\subsection{Speculative Decoding}
Speculative Decoding (SD) is a lossless method to accelerate LLM inference by using a faster draft mechanism, or drafter, to predict future tokens~\cite{leviathan2023fast}.
The LLM then validates these candidates in one forward pass in parallel, potentially accepting multiple tokens at once to reduce generation latency.
SD methods vary based on whether the approach requires a specific model (model-dependent vs. model-agnostic) and whether the drafter requires training (training-required vs. training-free).
Some SD approaches use an off-the-shelf smaller model in a model family as the drafter~\citep{xia2023speculative} and thus are model-dependent and training-free.
DistillSpec~\citep{zhou2024distillspec} trains a distilled drafter given any model, so the approach is model-agnostic but training-required.
Methods like EAGLE~\citep{li2024eagle} and MEDUSA~\citep{cai2024medusa} modify the LLM by adding auxiliary components, making them model-dependent and training-required.
Recently, SD methods that are both model-agnostic and training-free are favored for their plug-and-play convenience and broad applicability. They often leverage heuristics, small pre-trained models, or prompt and history information, such as lookahead strategies~\citep{zhao2024lookahead, fu2024break}, prompt-based lookups~\citep{saxena2023prompt}, or online draft construction~\citep{liu2024online}.

\section{\name Decoding}

We propose \name, a simple yet effective speculative decoding method that leverages cache tables to exploit the locality in language for speedup.
The table caches previously seen n-grams, divided into a \textit{leader} part and a \textit{follower} part (\S\ref{subsec:cache-structure}).
When queried with a leader, the table returns a list of followers that have previously appeared immediately after the leader.
At each decoding step, \name generates a tree of draft tokens by recursively querying the cache table and verifies them in parallel with one forward pass of the LLM (\S\ref{subsec:draft-generation-and-validation}).
Our strategy follows the intuition that n-grams which have recently appeared are likely to reappear.
After the LLM forward pass, \name updates the cache table to include new n-grams from accepted tokens, evicting stale entries with the least recently used (LRU) policy if necessary.
To avoid the cold-start problem, \name initializes the cache table with frequent n-grams observed in large training corpora (\S\ref{subsec:table-init}).
Despite being a simple method, \name achieves superior or comparable performance to many sophisticated model-agnostic and training-free methods developed in recent years (\S\ref{sec:experiment}).

We have open-sourced the implementation of \name with integration into the SpecBench benchmark suite.\footnote{\url{https://github.com/zyma98/Spec-Bench/tree/cacheback}}
We have also hosted the binary artifacts, i.e., frozen cache tables (\S\ref{subsec:table-init}), on Hugging Face for public access.\footnote{\url{https://huggingface.co/datasets/zyma98/cacheback_openwebtext_sample_100}}

\subsection{Cache Table Structure}
\label{subsec:cache-structure}

\begin{figure}
    \centering
    \includegraphics[width=0.482\textwidth]{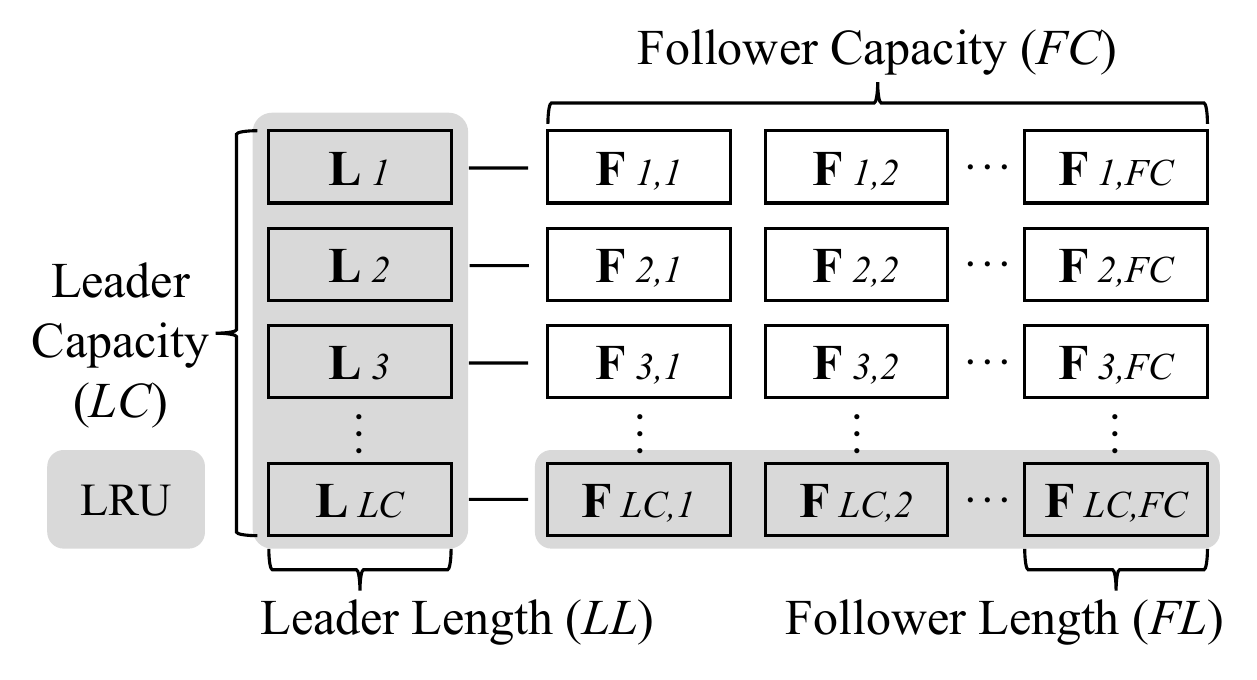}
    \caption{\name's cache table structure. Each leader is associated with a list of followers. Entries are evicted using the least recently used (LRU) policy.}
    \label{fig:table-structure}
\end{figure}

As shown in \autoref{fig:table-structure}, the cache table has a simple structure
in which the leaders and followers are both tuples of tokens (i.e., n-grams) and can be of different lengths, denoted by the leader length (\leadlen) and follower length (\folllen).
The table associates each leader with a list of followers.
When queried with a leader, the table returns the associated followers, or an empty list if the leader is absent.
The maximum number of leaders in a table and the maximum number of followers per leader are denoted by the leader capacity (\leadcap) and the follower capacity (\follcap), respectively.

The table updates its entries by accepting a leader-follower pair.
The table first checks if the leader exists in the table.
If not, the table creates a new entry for the leader and initializes the follower list with the new follower.
Otherwise, the table appends the new follower to the existing follower list if the follower is not already present.

When the number of leaders exceeds \leadcap, the table evicts the least recently used (LRU) leader and its followers upon inserting a new leader.
Both querying and inserting a leader update that leader's recency.
Likewise, if the number of followers of a leader exceeds \follcap, the table evicts the leader's LRU follower.
A notable difference is that the table updates a follower's recency only upon insertion.
Also, a follower compares its recency only against followers of the same leader.

The simplicity of the cache table structure enables fast lookup, insertion, and eviction.
For lookup, the table uses hash maps to locate leaders and followers.
For insertion and eviction, the table organizes leaders and followers in each LRU domain using doubly linked lists,
where items in a list are ordered by recency.
Therefore, the time complexity of lookup, insertion, and eviction operations is $O(1)$.

Standard libraries can readily support the construction of the table thanks to its simple structure.
Our prototype implementation simply uses the \texttt{OrderedDict} type provided by Python.

The memory consumption of the cache table can be estimated by counting the maximum number of tokens retained by the table, bounded by $O(\mathrm{LL} \cdot \mathrm{LC} \cdot \mathrm{FL} \cdot \mathrm{FC})$.
An ordered hash table like \texttt{OrderedDict} needs extra memory to maintain metadata including the bucket status and to store pointers in each linked element, but the asymptotic bound remains the same.

\subsection{Draft Generation and Validation}
\label{subsec:draft-generation-and-validation}

\begin{figure}
    \centering
    \includegraphics[width=0.48\textwidth]{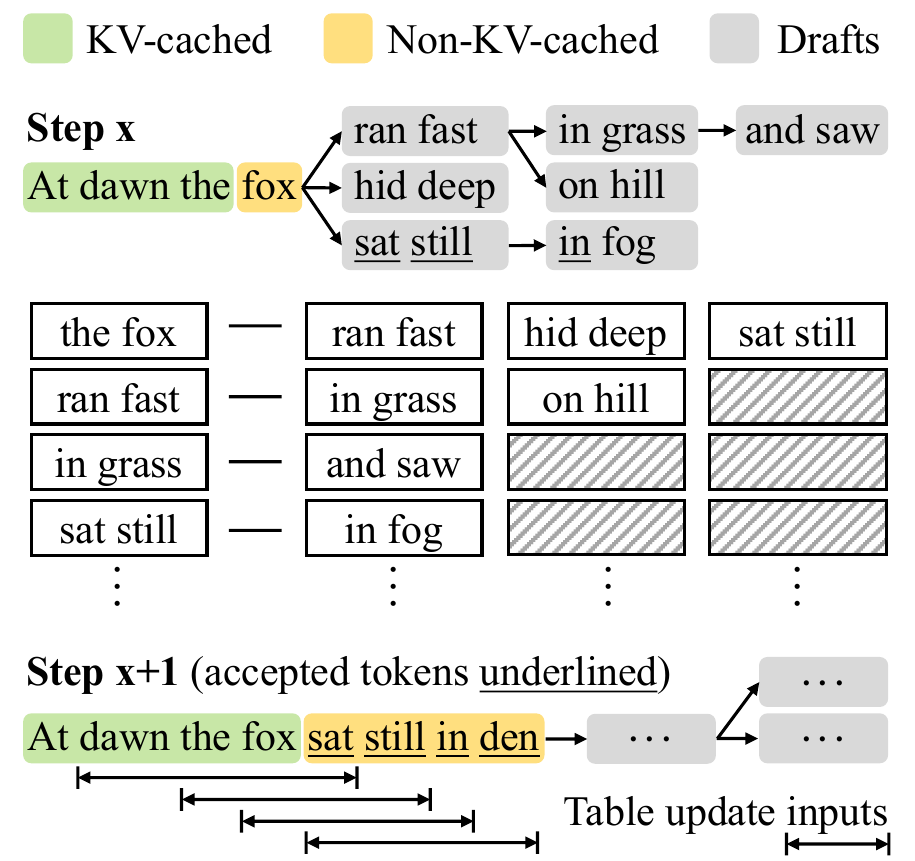}
    \caption{Overview of decoding steps. \name generates a draft tree by recursively querying the cache table and verifies it in one forward pass of the LLM using tree attention. In this example, the last draft branch except its last token is accepted. \name subsequently updates the cache table with the accepted tokens over a sliding window.}
    \label{fig:draft-tree}
\end{figure}

\name generates a tree of draft tokens by recursively querying the cache table.
\autoref{fig:draft-tree} shows an example of a draft tree generated for the prefix tokens ``At dawn the fox'' using a cache table where \leadlen = 2 and \folllen = 2.
We use a word to represent a token for illustrative purposes.
Tokens that are not KV-cached are those that were just accepted in the last decoding step.
In the first draft generation iteration, \name queries the table with the last two tokens in the sequence (``the fox'') and receives the list of followers containing ``ran fast,'' ``hid deep,'' and ``sat still.''
In subsequent iterations, \name attempts to grow the tree from its leaf nodes, querying the table with the last \leadlen tokens of the sequence when following the path from the root to a leaf node.
The growth of the draft token tree follows a breadth-first-search pattern.
Draft generation stops when either none of the leaf nodes has a follower in the cache table or the size of the tree reaches a predefined threshold.
We call this threshold the total draft length (\tdl), which counts the number of draft tokens plus the number of non-KV-cached tokens.

\name further introduces the chaining-reserved tokens (\crt) parameter to control the width versus depth of the tree.
\crt denotes the number of draft tokens reserved for the second or deeper level of the tree.
Without setting \crt, a draft tree can become wide enough that its first level exhausts \tdl.

\name employs tree attention to efficiently validate the draft tokens in one forward pass of the LLM.
\name builds a custom attention mask in which a token attends only to its ancestor tokens in the draft tree.
The LLM can then validate all branches of the tree in parallel as one input.
A custom GPU kernel may further improve the performance of \name.
We leave this as a future direction to explore.

An LLM forward pass always generates one more token in addition to the accepted tokens from the draft. Therefore, a decoding step generates one token in the worst case when it accepts no draft token, or one plus the longest branch length in the best case.

At the end of each decoding step, \name updates the cache table with a sliding window over the accepted tokens, as shown in \autoref{fig:draft-tree}.
The window captures all newly observed leader-follower pairs generated by the recent step.

\subsection{Table Initialization}
\label{subsec:table-init}

\name employs a dual-table approach to improve cold-start performance.
In addition to the dynamic cache table just described, \name prepares an additional \textit{frozen} table offline, filling it with frequent leader-follower pairs observed in large training corpora.
In each decoding step, \name first queries the dynamic table to form the draft tree and then the frozen one to further grow the tree if \tdl still allows.
The frozen table disregards insertions during decoding, and only the dynamic table is updated with accepted tokens.

Moreover, at the beginning of each decoding task, \name initializes the dynamic table with a sliding window over the prompt to populate it with the input context.

\section{Experiment}
\label{sec:experiment}

We conduct experiments on a desktop machine with an AMD Ryzen 5965WX CPU and four NVIDIA RTX 4090 GPUs.
We use one, two, and four GPUs to run the Vicuna 7B, 13B, and 33B models, respectively.

\begin{figure*}
    \centering
    \includegraphics[width=\textwidth]{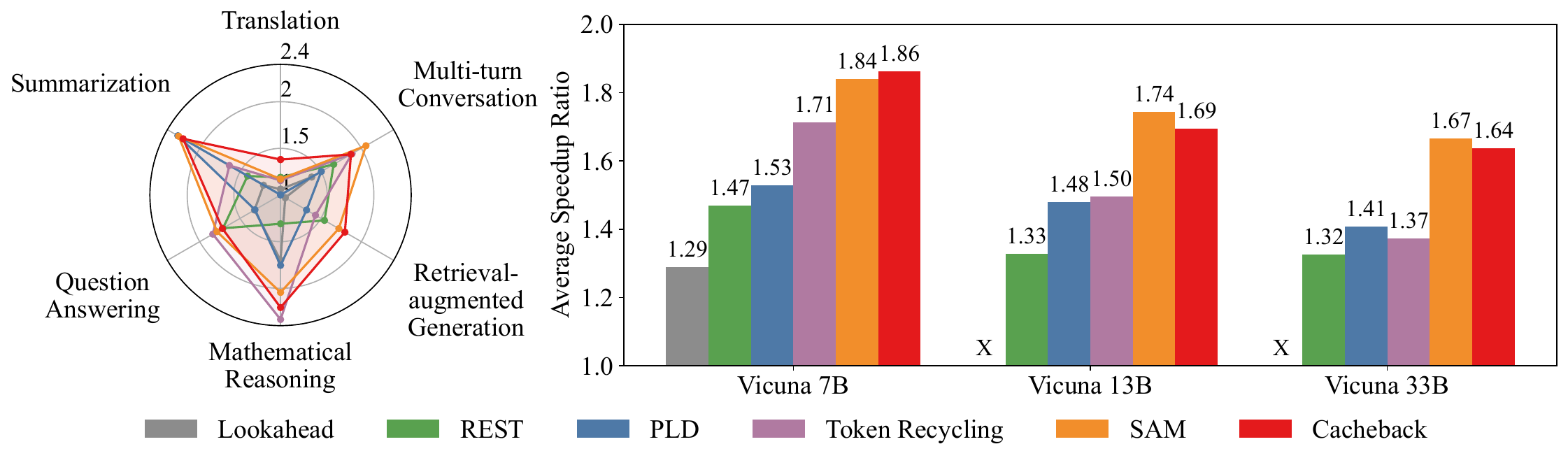}
    \vspace{-3ex}
    \caption{Wall-clock speedup ratio on SpecBench with Vicuna models. The radar plot shows the speedup on different task categories when running Vicuna 7B. \name achieves superior or comparable performance to other training-free model-agnostic methods.}
    \label{fig:radar}
\end{figure*}

To compare the performance of \name against other training-free model-agnostic methods, we use the SpecBench~\citep{xia2024unlocking} testing framework and dataset.
We modify SpecBench in two ways to ensure fairness among evaluated methods.
First, for stateful methods, including SAM Decoding~\citep{hu2024sam}, Token Recycling~\citep{luo2024turning}, and our approach, we reset the state object before running each test case.
Second, we build the static automaton as described in the SAM repository~\cite{sam-decoding} and include it when running SAM on SpecBench.
We also fix the SpecBench implementation of Retrieval-based Speculative Decoding (REST)~\citep{he2024rest} and Token Recycling so that the code can run with multiple GPUs.

For \name, we configure the cache table with \leadlen = 1, \leadcap = $\mathrm{2}^\mathrm{20}$, \folllen = 3, \follcap = 128, \tdl = 96, and \crt = 16.
We pick these values empirically for the best performance.
We configure \leadcap to be large to reduce the cache-miss rate and \follcap to be large to saturate \tdl.
With our configurations, a fully populated table uses at most a few GiB of DRAM, as analyzed in \S\ref{subsec:cache-structure}.

We build the frozen table by randomly sampling 1\% of the OpenWebText dataset~\citep{gokaslan2019openwebtext}.
The building procedure first picks the most frequent \leadcap n-grams of length \leadlen as the leaders in the table.
Then, for each leader, it selects the most frequent \follcap n-grams of length \folllen that appear after the leader in the dataset for inclusion in the follower list.

\begin{figure}
    \centering
    \includegraphics[width=0.40\textwidth]{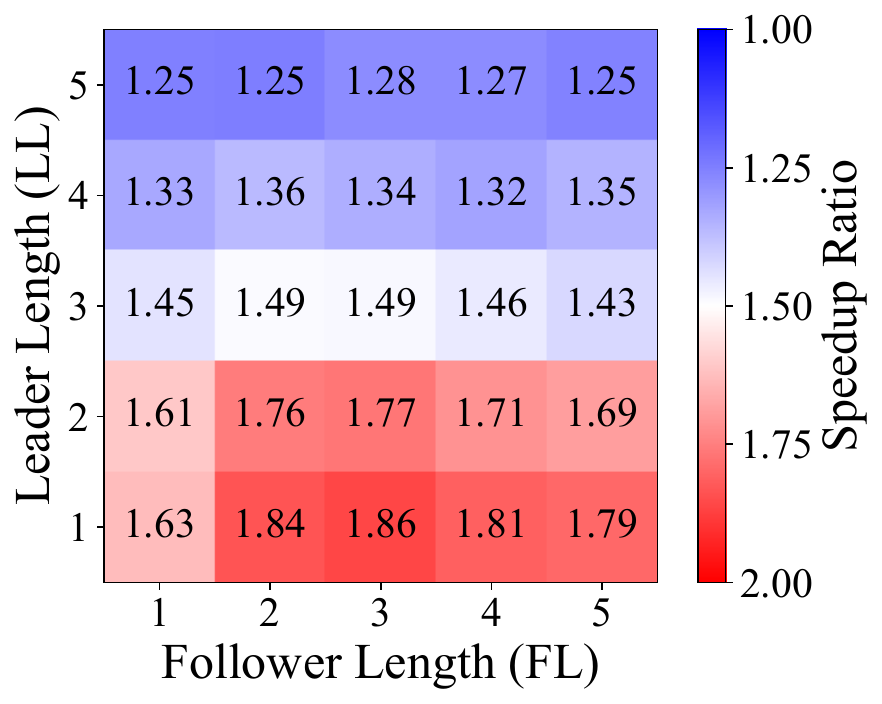}
    \caption{Speedup ratio of \name on SpecBench running Vicuna 7B with different \leadlen and \folllen settings.}
    \label{fig:heatmap}
\end{figure}

As shown in \autoref{fig:radar}, despite being simple, \name is on par with SAM Decoding based on suffix automata.
Furthermore, \name outperforms Prompt Lookup Decoding (PLD)~\citep{saxena2023prompt} that runs brute-force string matching, Lookahead Decoding (Lookahead)~\cite{fu2024break} that employs parallel Jacobi iteration, REST that leverages a database, and Token Recycling that constructs an adjacency matrix to generate drafts.
We note that the testing framework currently cannot run the Lookahead method with multiple GPUs.
Our results demonstrate that simpler methods can be just as effective as more sophisticated ones.

Notably, the translation task is particularly challenging for all evaluated SD methods.
This is partly because the generated words have very little relevance to the input context at the token level.
\name's lead in the translation domain demonstrates that our approach can effectively leverage language locality in the output text for speedup, suggesting its rapid adaptation to a new domain and its effectiveness for low-resource languages for which training a draft model is difficult.

We observe that the performance of \name exhibits an interesting pattern across different settings of \leadlen and \folllen, as shown in \autoref{fig:heatmap}, which plots the average speedup ratio of \name on SpecBench with Vicuna 7B while varying these two parameters.
In this figure, we set \leadcap, \follcap, \tdl, and \crt to the same values as before, but the trend remains the same with other configurations of these parameters.
\name consistently achieves the best performance when \leadlen is set to 1 and \folllen is around 3.
It may seem counterintuitive at first that \name runs fastest when \leadlen = 1.
However, \name's effectiveness is partly attributable to having multiple draft candidates in a tree, which increases the probability that some draft tokens are accepted.
With \leadlen = 1, the cache table can return more candidate followers with recent occurrences.
Meanwhile, \folllen = 3 strikes the best balance between the number of drafts and draft length.
A greater number of drafts increases the probability that at least some tokens from a draft will be accepted.
On the other hand, if a draft is very accurate, increasing draft length will result in more tokens being accepted in a step.

\begin{table}
    \centering
    \fontsize{10}{12}\selectfont
    \begin{tabular}{lrrr}
    \toprule
    Configuration  & Speedup         & MAT   & Token/s      \\ \midrule
    Dual           & 1.86$\times$    & 2.42  & 103.71        \\
    No Frozen      & 1.64$\times$    & 1.96  &  91.32        \\
    Only Frozen    & 1.28$\times$    & 1.59  &  68.11        \\
    \bottomrule
    \end{tabular}
    \caption{Speedup ratio, mean accepted tokens (MAT), and average token generation speed of \name running SpecBench with Vicuna 7B under different table configurations. The dual-table approach is necessary for good performance.}
    \label{tab:frozen-contrib}
\end{table}

Finally, we demonstrate in \autoref{tab:frozen-contrib} that the dual-table approach is essential for \name to achieve good performance.
Without the frozen table, both the mean accepted tokens (MAT) and the speedup ratio drop significantly due to the cold-start problem.
Moreover, since the frozen table cannot reflect the specific context of a decoding loop, using it alone is also suboptimal.

\section{Conclusion}

We propose \fullname, a simple yet effective speculative decoding method that leverages LRU cache tables to exploit the locality in language to accelerate LLM inference.
Our results show that \name achieves superior or comparable performance to many sophisticated training-free model-agnostic methods.
Due to its simplicity, \name can be easily integrated into existing LLM frameworks.
Moreover, because \name organizes draft tokens as a tree, it can be combined with other SD methods for further speedup by inserting their predicted drafts as additional branches, similar to the combination of SAM decoding and EAGLE~\citep{hu2024sam}.
Finally, with \name's inherited strengths from CLMs, our approach exhibits rapid adaptation to specific domains, as evidenced by its strong performance in translation.

\section*{Acknowledgments}
This work was supported in part by National Science Foundation (NSF) Athena AI Institute under Award 2112562. The authors are grateful for the useful feedback from their reviewers.

\newpage

\section*{Limitations}

Our exploration of \name has several remaining areas for investigation.
The impact of different corpora for initializing the frozen table remains unstudied, as does performance variation across different GPU architectures and LLM models.
Additionally, our evaluation is currently limited to the SpecBench dataset, which may not represent all possible use cases.
Moreover, the performance of \name exhibits sensitivity to configuration parameters, with optimal settings likely varying across different tasks, language models, and hardware configurations.
A theoretical analysis of these parameters is still lacking, and developing an automatic parameter tuning method would be valuable.
We leave these investigations for future work.

\bibliography{abr-long,references}

@string{icml="Proc. Int. Conf. Machine Learning (ICML)"}

@string{iclr="Proc. Int. Conf. Learning Representations (ICLR)"}

@inproceedings{fu2024break,
    title={Break the sequential dependency of LLM inference using LOOKAHEAD DECODING},
    author={Fu, Yichao and Bailis, Peter and Stoica, Ion and Zhang, Hao},
    booktitle=ICML,
    year={2024},
    url = {https://openreview.net/forum?id=eDjvSFOkXw},
}

@misc{saxena2023prompt,
    title = {Prompt Lookup Decoding},
    author = {Apoorv Saxena},
    year = {2023},
    howpublished = {https://github.com/apoorvumang/prompt-lookup-decoding/},
    url = {https://github.com/apoorvumang/prompt-lookup-decoding/}
}

@inproceedings{zhou2024distillspec,
    title={{DistillSpec}: Improving Speculative Decoding via Knowledge Distillation},
    author={Yongchao Zhou and Kaifeng Lyu and Ankit Singh Rawat and Aditya Krishna Menon and Afshin Rostamizadeh and Sanjiv Kumar and Jean-Fran{\c{c}}ois Kagy and Rishabh Agarwal},
    booktitle=ICLR,
    year={2024},
    url = {https://openreview.net/forum?id=rsY6J3ZaTF}
}

@inproceedings{li2024eagle,
    title={EAGLE: speculative sampling requires rethinking feature uncertainty},
    author={Li, Yuhui and Wei, Fangyun and Zhang, Chao and Zhang, Hongyang},
    booktitle=icml,
    year={2024},
    url = {https://openreview.net/forum?id=1NdN7eXyb4}
}

@inproceedings{cai2024medusa,
    title={{MEDUSA}: Simple {LLM} inference acceleration framework with multiple decoding heads},
    author={Cai, Tianle and Li, Yuhong and Geng, Zhengyang and Peng, Hongwu and Lee, Jason D and Chen, Deming and Dao, Tri},
    booktitle=ICML,
    year={2024},
    url = {https://openreview.net/forum?id=PEpbUobfJv}
}

@inproceedings{liu2024online,
    title={Online speculative decoding},
    author={Liu, Xiaoxuan and Hu, Lanxiang and Bailis, Peter and Cheung, Alvin and Deng, Zhijie and Stoica, Ion and Zhang, Hao},
    booktitle=icml,
    year={2024},
    url = {https://openreview.net/forum?id=BPQHXwVNvl}
}

@inproceedings{he2024rest,
    title={{REST}: Retrieval-Based Speculative Decoding},
    author={He, Zhenyu and Zhong, Zexuan and Cai, Tianle and Lee, Jason and He, Di},
    booktitle={Proc. Conf. North American Chapter of the Association for Computational Linguistics: Human Language Technologies},
    year={2024},
    doi = {10.18653/v1/2024.naacl-long.88}
}

@inproceedings{xia2023speculative,
    title={Speculative Decoding: Exploiting Speculative Execution for Accelerating Seq2seq Generation},
    author={Xia, Heming and Ge, Tao and Wang, Peiyi and Chen, Si-Qing and Wei, Furu and Sui, Zhifang},
    booktitle={Findings of the Association for Computational Linguistics: EMNLP},
    year={2023},
    url = {https://proceedings.neurips.cc/paper_files/paper/2024/file/9cb5b083ba4f5ca6bd05dd307a2fb354-Paper-Conference.pdf}
}

@article{hu2024sam,
    title={{SAM} Decoding: Speculative Decoding via Suffix Automaton},
    author={Hu, Yuxuan and Wang, Ke and Zhang, Xiaokang and Zhang, Fanjin and Li, Cuiping and Chen, Hong and Zhang, Jing},
    journal={arXiv preprint arXiv:2411.10666},
    year={2024},
    url={https://arxiv.org/abs/2411.10666},
}

@misc{sam-decoding,
    title = {Official Implementation of {SAM-Decoding}: Speculative Decoding via Suffix Automaton},
    author = {Yuxuan Hu},
    year= {2024},
    howpublished = {https://github.com/hyx1999/SAM-Decoding},
    url = {https://github.com/hyx1999/SAM-Decoding}
}

@inproceedings{zhao2024lookahead,
    title={Lookahead: An inference acceleration framework for large language model with lossless generation accuracy},
    author={Zhao, Yao and Xie, Zhitian and Liang, Chen and Zhuang, Chenyi and Gu, Jinjie},
    booktitle={Proc. ACM SIGKDD Conf. Knowledge Discovery and Data Mining (KDD)},
    year={2024},
    doi={10.1145/3637528.3671614},
}

@article{luo2024turning,
    title={Turning Trash into Treasure: Accelerating Inference of Large Language Models with Token Recycling},
    author={Luo, Xianzhen and Wang, Yixuan and Zhu, Qingfu and Zhang, Zhiming and Zhang, Xuanyu and Yang, Qing and Xu, Dongliang and Che, Wanxiang},
    journal={arXiv preprint arXiv:2408.08696},
    year={2024},
    url={https://arxiv.org/abs/2408.08696}
}

@misc{gokaslan2019openwebtext,
    title={OpenWebText Corpus},
    author={Gokaslan, Aaron and Cohen, Vanya and Pavlick, Ellie and Tellex, Stefanie},
    howpublished={\url{http://Skylion007.github.io/OpenWebTextCorpus}},
    year={2019}
}

@inproceedings{xia2024unlocking,
  title={Unlocking Efficiency in Large Language Model Inference: A Comprehensive Survey of Speculative Decoding},
  author={Xia, Heming and Yang, Zhe and Dong, Qingxiu and Wang, Peiyi and Li, Yongqi and Ge, Tao and Liu, Tianyu and Li, Wenjie and Sui, Zhifang},
  booktitle={Findings of Association for Computational Linguistics},
  year={2024},
  doi = "10.18653/v1/2024.findings-acl.456",
}

@article{kuhn1990cache,
  title={A cache-based natural language model for speech recognition},
  author={Kuhn, Roland and De Mori, Renato},
  journal={IEEE Trans. Pattern Analysis and Machine Intelligence},
  year={1990},
  doi={10.1109/34.56193}
}

@inproceedings{leviathan2023fast,
  title={Fast inference from transformers via speculative decoding},
  author={Leviathan, Yaniv and Kalman, Matan and Matias, Yossi},
  booktitle=ICML,
  year={2023},
  url={https://openreview.net/forum?id=C9NEblP8vS}
}

@article{futrell2015large,
  title={Large-scale evidence of dependency length minimization in 37 languages},
  author={Futrell, Richard and Mahowald, Kyle and Gibson, Edward},
  journal={Proc. National Academy of Sciences},
  year={2015},
  doi={10.1073/pnas.1502134112}
}

@inproceedings{jelinek1991dynamic,
  title={A dynamic language model for speech recognition},
  author={Jelinek, Frederick and Merialdo, Bernard and Roukos, Salim and Strauss, Martin},
  booktitle={Proc. Wrkshp. Speech and Natural Language},
  year={1991},
  url={https://aclanthology.org/H91-1057.pdf}
}

@inproceedings{clarkson1997language,
  title={Language model adaptation using mixtures and an exponentially decaying cache},
  author={Clarkson, Philip R and Robinson, Anthony J},
  booktitle={Proc. IEEE Int. Conf. Acoustics, Speech, and Signal Processing (ICASSP)},
  year={1997},
  doi = {10.1109/ICASSP.1997.596049}
}

\end{document}